\documentclass[10pt,twocolumn,letterpaper]{article}

\usepackage{iccv}
\usepackage{times}
\usepackage{epsfig}
\usepackage{graphicx}
\usepackage{amsmath}
\usepackage{amssymb}
\usepackage{subfigure}
\usepackage{color}
\usepackage{comment}
\usepackage{here}

\usepackage[breaklinks=true,bookmarks=false]{hyperref}

\iccvfinalcopy 

\def\iccvPaperID{****} 

\begin{document}

\title{Changing Fashion Cultures}

\author{Kaori Abe$^*$, Teppei Suzuki$^*$, Shunya Ueta, Yutaka Satoh, Hirokatsu Kataoka\\
National Institute of Advanced Industrial Science and Technology (AIST)\\
Tsukuba, Ibaraki, Japan\\
{\tt\small \{abe.keroko, suzuki-teppei, shunya.ueta, yu.satou, hirokatsu.kataoka\}@aist.go.jp}
\and
Akio Nakamura\\
Tokyo Denki University\\
Adachi, Tokyo, Japan\\
{\tt\small nkmr-a@cck.dendai.ac.jp}
}

\maketitle

\begin{abstract}
The paper presents a novel concept that analyzes and visualizes worldwide fashion trends. Our goal is to reveal cutting-edge fashion trends without displaying an ordinary fashion style. To achieve the fashion-based analysis, we created a new fashion culture database (FCDB), which consists of 76 million geo-tagged images in 16 cosmopolitan cities. By grasping a fashion trend of mixed fashion styles, the paper also proposes an unsupervised fashion trend descriptor (FTD) using a fashion descriptor, a codeword vector, and temporal analysis. To unveil fashion trends in the FCDB, the temporal analysis in FTD effectively emphasizes consecutive features between two different times. In experiments, we clearly show the analysis of fashion trends and fashion-based city similarity. As the result of large-scale data collection and an unsupervised analyzer, the proposed approach achieves world-level fashion visualization in a time series. The code, model, and FCDB will be publicly available after the construction of the project page.
\end{abstract}

\begin{figure*}[t]
\centering
\subfigure[Fashion trend changes in New York, Paris, London, and Tokyo, 2014--2016. Our goal is to reveal the latest trends for each year.]{\includegraphics[width=0.35\linewidth]{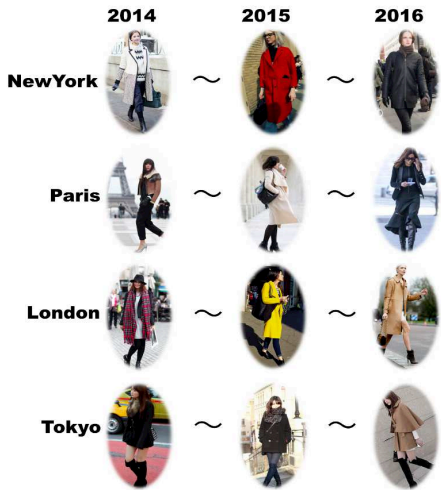}
\label{fig:trends}}
\subfigure[Fashion culture database (FCDB; ours)]{\includegraphics[width=0.63\linewidth]{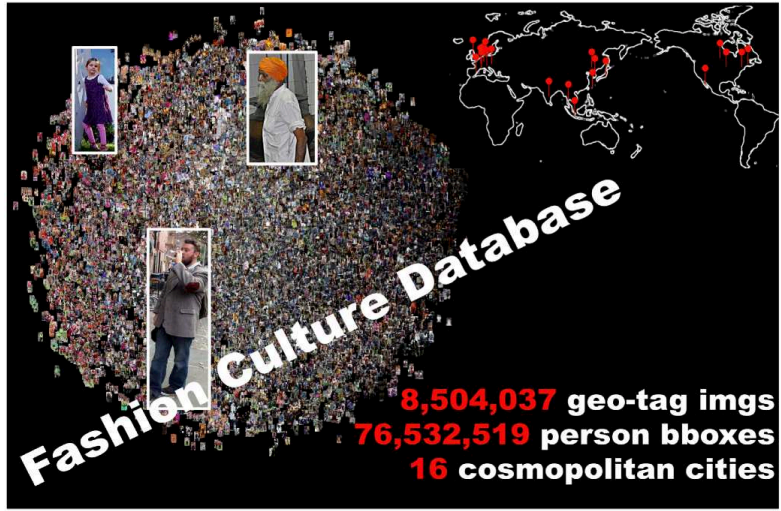}
\label{fig:fcdb_abst}}
\caption{Worldwide fashion change analysis from the self-collected fashion culture database (FCDB). We analyze changing fashion cultures for every season (a). The images are taken from Flickr, under a Creative Commons license. The FCDB consists of 8M geo-tagged images and 76M person bounding boxes in 16 cosmopolitan cities (b).}
\label{fig:concept}
\end{figure*}

\section{Introduction}

\renewcommand{\thefootnote}{\fnsymbol{footnote}}
\footnote[0]{$^*$ Indicates equal contribution.}
\renewcommand{\thefootnote}{\arabic{footnote}}

Farnand Braudel stated that the fashion history started to continuously change in the 14th century~\cite{Braudel1981}. The early fashion change was mainly seen in European cultures. Through the mass production of the industrial revolution, the numbers of various fashion styles and designers have simultaneously increased all over the world. Due to fashion cultures in the 20th century, fashion trends have been changing each year. Especially, sophisticated fashion communities that largely affect world trends are gathered in cosmopolitan cities such as New York and Paris. Undoubtedly, fashion trends are affected by local culture and climate change; however, these fashion communities tend to lead the world fashion industry. Figure~\ref{fig:trends} shows examples of fashion changes in New York, Paris, London, and Tokyo, 2014--2016. The cutting edge of fashion styles seems to shift every year.

Moreover, photo-sharing websites such as Flickr and Instagram exploded the number of pictures. The enormous amount of images helps to characterize fashion trends. Thus, worldwide data collection from image locations and times highly accelerates large-scale spatio-temporal fashion trends. Moreover, digital fashion analysis has been studied in the field of computer vision (e.g., cloth parsing~\cite{YamaguchiCVPR2012}, style recognition~\cite{KiapourECCV2014}). A more recent study treats fashion recognition with compact convolutional neural network (CNN) activation~\cite{SimoSerraCVPR2016}. The author's experiments showed that a style descriptor is robust to image background. That positive result allows us to analyze a fashion style in the real world.

In this paper, we propose the concept of ``changing fashion cultures", which analyzes and visualizes worldwide fashion trends in a time series. 
We highlight our contributions below:

\textbf{\underline{Conceptual contribution}:} We propose a novel concept to analyze and visualize worldwide fashion trends. We refer to this concept as changing fashion cultures, which enable the unveiling of the latest fashion styles. To the best of our knowledge, this is the first study of vision-based temporal fashion trends analysis by a geo-tagged image database. To realize the concept, we created a new fashion database that contains over 76M fashion snaps  in 16 cosmopolitan cities (see Figure~\ref{fig:fcdb_abst}).

\textbf{\underline{Technical contribution}:} An unsupervised fashion trend descriptor (FTD) created from a feature description, clustering, a codeword vector, and temporal analysis is also proposed in this paper. The FTD is calculated by two consecutive times to appear the latest fashion styles. To improve the codeword-based representation, we experimentally enhance the style descriptor with an extra feature. The FTD is a simple codeword-based approach, but it is an effective unsupervised analyzer.

\textbf{\underline{Experimental contribution}:} In experiments, we clearly show our analysis of fashion trends and fashion-based city similarity. As the result of large-scale data collection and the unsupervised analyzer, the proposed approach enables us to realize world-level fashion visualization in a time series.

\section{Related work}
\subsection{Fashion database}

The first notable work in fashion analysis was cloth parsing, which used a graphical model~\cite{YamaguchiCVPR2012}. The resulting Fashionista dataset  was released to the research community, and the database has induced the recent flow of fashion analyses.

Some studies have focused on a fashion attribute analysis~\cite{ChenECCV2012,YamaguchiBMVC2015,LiuCVPR2016}. As an example, the DeepFashion dataset was released in the age of big data~\cite{LiuCVPR2016}. This large-scale database strengthened attribute estimation through a data-driven deep learned architecture. Liu \textit{et al.} verified that fashion landmarks improve the attribute recognition in a fashion database~\cite{LiuECCV2016}.

In style classification, the Hipster Wars dataset~\cite{KiapourECCV2014} provided a source of fashion style recognition, which is a difficult problem, because fashion style is defined by high-level features and not only low-level features such as color, texture, and oriented edges. In their experiment, a high-level feature could effectively divide fashion styles in the Hipster Wars dataset. Similar data come from the Fashionability dataset~\cite{SimoSerraCVPR2015}. The Fashionability dataset contains 144K images cropped  with fashion snaps and geo-tags in the real world. A style descriptor~\cite{SimoSerraCVPR2016} is deeply learned in the framework of weak supervision in the Fashionability dataset. The style descriptor calculates a compact feature by treating a similar fashion pair located at near coordinates in Euclidean space. We believe that the compact and accurate representation allows us to conduct very large-scale fashion analyses.

To explicitly observe worldwide fashion trends, we have collected an enormous number of geo-tagged images from photo-sharing websites. Our database is about 100 times larger than the other fashion databases (see Table~\ref{tab:fashiondbs}).

\begin{table*}[t]
\begin{center}
\caption{Representative fashion databases}
\begin{tabular}{llrccccc}
\hline
\multicolumn{1}{l}{Database} & \multicolumn{1}{l}{Task} & \multicolumn{1}{r}{\#\ images} & \multicolumn{1}{c}{\#\ category} & \multicolumn{1}{c}{Public} & \multicolumn{1}{c}{Geo-tag} & \multicolumn{1}{c}{PersonTag} & \multicolumn{1}{c}{TimeStamp} \\
\hline \hline
\multicolumn{1}{l}{HipsterWars~\cite{KiapourECCV2014}} & \multicolumn{1}{l}{Style classification} & \multicolumn{1}{r}{1,893} & \multicolumn{1}{r}{5} & \multicolumn{1}{c}{\checkmark} & \multicolumn{1}{c}{} & \multicolumn{1}{c}{} & \multicolumn{1}{c}{} \\
\multicolumn{1}{l}{Fashionista~\cite{YamaguchiCVPR2012}} & \multicolumn{1}{l}{Parsing, pose estimation} & \multicolumn{1}{r}{158,235} & \multicolumn{1}{r}{53} & \multicolumn{1}{c}{\checkmark} & \multicolumn{1}{c}{} & \multicolumn{1}{c}{\checkmark} & \multicolumn{1}{c}{} \\
\multicolumn{1}{l}{Fashion144k~\cite{SimoSerraCVPR2015}} & \multicolumn{1}{l}{Style classification} & \multicolumn{1}{r}{144,169} & \multicolumn{1}{r}{N/A} & \multicolumn{1}{c}{\checkmark} & \multicolumn{1}{c}{\checkmark} & \multicolumn{1}{c}{} & \multicolumn{1}{c}{} \\
\multicolumn{1}{l}{DeepFashion~\cite{LiuCVPR2016}} & \multicolumn{1}{l}{Attribute estimation} & \multicolumn{1}{r}{800,000} & \multicolumn{1}{r}{1,050} & \multicolumn{1}{c}{\checkmark} & \multicolumn{1}{c}{} & \multicolumn{1}{c}{\checkmark} & \multicolumn{1}{c}{} \\
\multicolumn{1}{l}{FCDB (ours)} & \multicolumn{1}{l}{Fashion trend analysis} & \multicolumn{1}{r}{\textbf{76,532,519}} & \multicolumn{1}{r}{16} & \multicolumn{1}{c}{\checkmark} & \multicolumn{1}{c}{\checkmark} & \multicolumn{1}{c}{\checkmark} & \multicolumn{1}{c}{\checkmark} \\
\hline
\end{tabular}
\label{tab:fashiondbs}
\end{center}
\end{table*}

\subsection{Geo-tagged image analysis}

Studies using geo-tagged images collected from photo-sharing websites have influenced the computer vision field in the last decade. Flickr and Panoramio have supplied a developer tool to promote their new web service. Along the same lines, the Yahoo! Creative Commons 100M Database (YFCC100M)~\cite{ThomeeACM2016} was created, and it consists of 100 million images and geo-tagged information. This data collection leads the world-level 3D reconstruction of several landmarks~\cite{HeinlyCVPR2015}. Other studies have demonstrated landmark recognition from a picture taken at some location~\cite{HaysCVPR2008} and database construction for landmark retrieval~\cite{CrandallWWW2009}.

Our paper is highly inspired by city identification via attribute analysis~\cite{ZhouECCV2014,LiuGeoJournal2016}. This successful identification derived the characteristics of 21 selected cities over 3 continents. The scene attributer and the SUN database was used to describe an identity based on 7 important attributes (e.g., water coverage, green space coverage). The difference between our analysis and that by \cite{ZhouECCV2014} is shown below:
\begin{itemize}
  \item \textbf{City identity}: Due to scene-based recognition, their approach tends to be a static analysis.
  \item \textbf{Changing fashion cultures (ours)}: Fashion trends dramatically shift depending on the intention of designers and fashion industries. Therefore, we must include the temporal representation in geo-tagged analysis. That is, we must visualize the fashion trends in multiple cities.
\end{itemize}

\section{Fashion culture database (FCDB)}

In this section, we summarize the FCDB and discuss a scenario for worldwide fashion analysis, database collection, and annotation.

\subsection{Summary of FCDB}

We generate the fashion culture database (FCDB), which helps to achieve a temporal analysis for worldwide fashion changes in individual global cities.

\begin{figure}[t] 
\centering
 \includegraphics[width=1.0\linewidth]{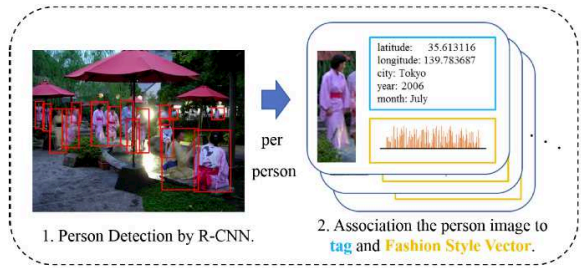}
 \caption{Construction of the FCDB: For each city, the database contains downloaded images, person-cropped images, tag information (tag), and feature vectors (vec). The person is detected and cropped with the Faster R-CNN. The tag information is based on the time stamp and geo-location. The combined StyleNet and Lab  are experimentally selected as feature vectors.}
\label{fig:dbconstruction}
\end{figure}

The FCDB is collected from the Yahoo! Creative Commons 100M Database (YFCC100M)~\cite{ThomeeACM2016}, which contains 100 million Flickr images. We focus on 21 global cities based on city identity~\cite{ZhouECCV2014}. However, we exclude the cosmopolitan cities if the number of images is less than 100K. Consequently, 16 of the 21 cosmopolitan cities remain. To analyze temporal fashion culture changes, we insert a time stamp at each image from 2000 to 2015. Table~\ref{tab:fcdb} shows the list of 16 cities and their longitudes and latitudes. The FCDB includes a total of 76,532,519 images. To the best of our knowledge, this is the largest existing fashion database (see Table~\ref{tab:fashiondbs}). The FCDB enables global-scale fashion trend visualization. We also crop to person-centered patches with the human detection method Faster R-CNN~\cite{RenNIPS2015} (see Figure~\ref{fig:dbconstruction}). 

\begin{table}[t]
\centering
\caption{16 cities and coordinates in the FCDB}
\begin{tabular}{lcc}
\hline
\multicolumn{1}{l}{City} & \multicolumn{1}{c}{Longitude} & \multicolumn{1}{c}{Latitude} \\
\hline \hline
\multicolumn{1}{l}{London} & \multicolumn{1}{c}{-0.12776} & \multicolumn{1}{c}{51.50735} \\
\multicolumn{1}{l}{New York} & \multicolumn{1}{c}{-74.0059} & \multicolumn{1}{c}{40.71278} \\
\multicolumn{1}{l}{Boston} & \multicolumn{1}{c}{-71.0589} & \multicolumn{1}{c}{42.36008} \\
\multicolumn{1}{l}{Paris} & \multicolumn{1}{c}{2.352222} & \multicolumn{1}{c}{48.85661} \\
\multicolumn{1}{l}{Toronto} & \multicolumn{1}{c}{-79.3832} & \multicolumn{1}{c}{43.65323} \\
\multicolumn{1}{l}{Barcelona} & \multicolumn{1}{c}{2.173403} & \multicolumn{1}{c}{41.38506} \\
\multicolumn{1}{l}{Tokyo} & \multicolumn{1}{c}{139.6917} & \multicolumn{1}{c}{35.68949} \\
\multicolumn{1}{l}{San Francisco} & \multicolumn{1}{c}{-122.419} & \multicolumn{1}{c}{37.77493} \\
\multicolumn{1}{l}{Hong Kong} & \multicolumn{1}{c}{114.1095} & \multicolumn{1}{c}{22.39643} \\
\multicolumn{1}{l}{Zurich} & \multicolumn{1}{c}{8.541694} & \multicolumn{1}{c}{47.37689} \\
\multicolumn{1}{l}{Seoul} & \multicolumn{1}{c}{126.978} & \multicolumn{1}{c}{37.56654} \\
\multicolumn{1}{l}{Beijing} & \multicolumn{1}{c}{116.4074} & \multicolumn{1}{c}{39.90421} \\
\multicolumn{1}{l}{Bangkok} & \multicolumn{1}{c}{100.5018} & \multicolumn{1}{c}{13.75633} \\
\multicolumn{1}{l}{Singapore} & \multicolumn{1}{c}{103.8198} & \multicolumn{1}{c}{1.352083} \\
\multicolumn{1}{l}{Kuala Lumpur} & \multicolumn{1}{c}{101.6869} & \multicolumn{1}{c}{3.139003} \\
\multicolumn{1}{l}{New Delhi} & \multicolumn{1}{c}{77.20902} & \multicolumn{1}{c}{28.61394} \\
\hline
\end{tabular}
\label{tab:fcdb}
\end{table}

\subsection{Two tasks for worldwide fashion analysis}

Based on fashion styles, we determined two different tasks: spatial city perception and temporal fashion trend analysis. In this paper, we define the fashion style of clothes according to social status, climate, culture, and religious and personal preference. Moreover, we treat similar appearances of style (e.g., usage of a style descriptor~\cite{SimoSerraCVPR2016}) as a fashion style to efficiently represent fashion styles in the extremely large FCDB. To be exact, unsupervised clustering is used to handle countless fashion styles as a limited number.

We solve the spatial and the temporal fashion analyses on the FCDB. The two divided tasks are defined as follows:

\begin{itemize}
\item City perception \& similarity as a spatial fashion analysis: We spatially analyze the 16 cities based on \cite{ZhouECCV2014}. First, we try to identify a city with only a fashion-based descriptor in~\ref{sec:cityperceptions}. A second spatial analysis shows a city similarity graph at a particular time in~\ref{sec:citysimilarity}. The similarity graph is calculated with a codeword vector that represents the culmination of fashion styles.
\item Fashion trend as a temporal fashion analysis: We analyze temporal subtraction between two consecutive times (e.g., years $y$ \& $y+1$) in order to show the latest fashion trend at the moment. The task of temporal fashion analysis must visualize several fashion trends as examples. The visualization is shown in~\ref{sec:fashiontrend}.
\end{itemize}

\begin{figure*}[t]
\centering
\subfigure[Number of collected geo-tagged images on the FCDB: ``Image" and ``human image by R-CNN" shows the captured geo-tagged images from the YFCC100M and person-centered cropped images]{\includegraphics[width=0.36\linewidth]{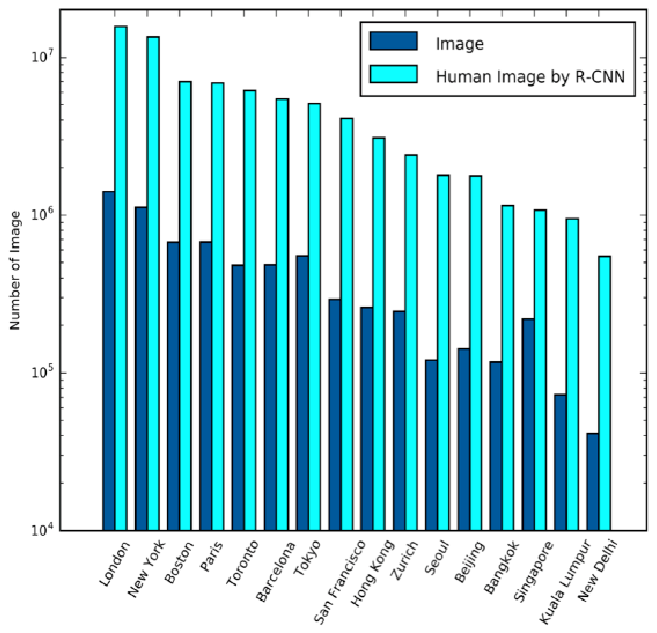}
\label{fig:number_of_image}}
\subfigure[Number of collected images per year]{\includegraphics[width=0.40\linewidth]{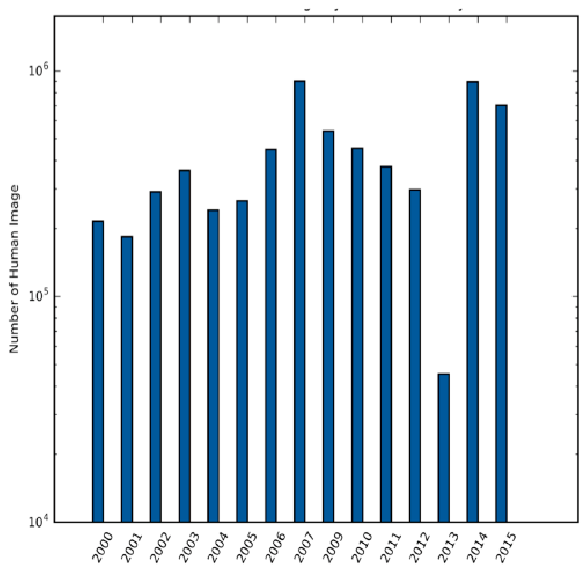}
\label{fig:numofimage_year}}
\caption{Number of image}
\label{fig:numofimage}
\end{figure*}

\subsection{Image collection and automatic annotation}

To collect a ten-million-order fashion database, we capture geo-tagged images from Flickr. As shown in Figure~\ref{fig:dbconstruction},  the FCDB contains (i) captured images from the YFCC100M dataset, (ii) person-cropped images, and (iii) time stamps and geo-tag information. We set a 100 km radius around a city in order to collect images (i) based on longitude and latitude (see Table~\ref{tab:fcdb}). The 16 cities in the FCDB are \{London, New York, Boston, Paris, Toronto, Barcelona, Tokyo, San Francisco, Hong Kong, Zurich, Seoul, Beijing, Bangkok, Singapore, Kuala Lumpur, New Delhi\}. The areas do not overlap each other. To create the images shown in (ii), we apply VGG16-model Faster R-CNN~\cite{RenNIPS2015}, which is a real-time object detection framework. We set the thresholding value as 0.8 and use only the person-label in the Pascal VOC pre-trained model. The person-cropped image allows us to eliminate noises in a background. A set of geo-tag and time stamp in (iii) is replicated from the YFCC100M dataset. Figures~\ref{fig:number_of_image} and \ref{fig:numofimage_year}  indicate the numbers of images collected from the YFCC100M and cropped human patches, respectively. 

The cities are annotated with longitudes and latitudes in Table~\ref{tab:fcdb}. Therefore, we can automatically execute the annotations on the FCDB by positional information. The database have been cross
validated by the annotators and extra validators.

\section{Fashion Style Features}

\subsection{Local feature descriptor for fashion analysis}
We calculate StyleNet as a local feature descriptor. Although StyleNet is strong for representing a fashion style, we believe that an extra feature is beneficial for complementing StyleNet in a cropped-person image. Then we experimentally assign L*a*b as a color feature.

StyleNet is based on the concept of deep similarity, which directly calculates the distance between two fashion snaps in Euclidean space by using deep learning~\cite{SimoSerraCVPR2016}. The goal is to learn the similarity of fashion style as a Euclidean distance. The squared distance $D(.,.)$ between two images $I_1$ and $I_2$ in Euclidean space is as follows:
\begin{eqnarray}
D(f(I_1), f(I_2)) = ||f(I_1) - f(I_2)||_{2}^{2}
\end{eqnarray}
where  $f(.)$ is a projection function  from the image space into the Euclidean space which represents fashion similarity. This projection function is trained with a deep learning architecture. It is jointly trained with cross-entropy and ranking loss with triplet $\tau = (I_{-}, I, I_{+})$ that contains anchor $I$, positive $I$ ($I_{+}$; similar image to $I$), and negative $I$ ($I_{-}$; dissimilar image to $I$) images. 


Here we employ the Fashion144k dataset~\cite{SimoSerraCVPR2015} to train the StyleNet descriptor. The Fashion144k dataset is collected from a web-based image. The characteristics are very similar to the FCDB, and so we do not need to apply additional training to StyleNet. Given an image $I$, the pre-trained StyleNet outputs a 128-dim vector in Euclidean space. Similar fashion styles tend to be close to each other in the Euclidean space. However, we believe that a more detailed color feature is required in the street style downloaded by Flickr. Therefore, the combined feature of StyleNet and L*a*b is assigned.

We use the L*a*b feature to improve a StyleNet feature. In our implementation, the L*a*b feature originally has 256 dimensions and is compressed into 128 dimensions by principal component analysis (PCA). To concatenate L*a*b and the StyleNet feature, we expect to get color robustness. We show the experiment of feature improvement in Section~\ref{sec:featureimprovement}.

\begin{figure*}[t]
\centering
 \includegraphics[width=1.0\linewidth]{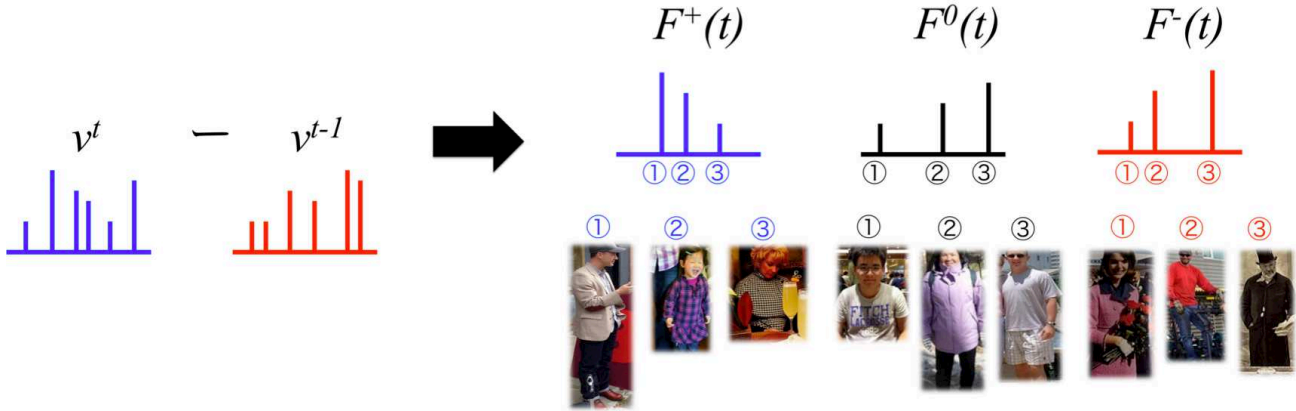}
 \caption{Fashion trend descriptor (FTD): FTD ($F$) is the subtraction result between two different codeword vectors $v^{t}$ and $v^{t-1}$. The FTD ($F$) consists of $F^{+}(t)$, $F^{0}(t)$, and $F^{-}(t)$. $F^{+}(t)$ and $F^{-}(t)$ are appearing and disappearing fashion styles at time $t$, respectively. Each bin in $F$ corresponds to a fashion style based on the BoW. This figure includes nine fashion examples at each FTD ($F$).}
\label{fig:ftd}
\end{figure*}

\subsection{Fashion Trend Descriptor (FTD)}
To unveil the latest fashion trends without displaying an ordinary fashion style, we propose the fashion trend descriptor (FTD) $F$, which is the subtraction result between two codeword vectors for two consecutive times. The combined vector of StyleNet+L*a*b is clustered to construct a bag-of-words (BoW)~\cite{CsurkaECCVW2004} as a codeword vector. To visualize a fashion trend as a temporal fashion analysis, we employ the FTD. The detailed description is shown in Figure~\ref{fig:ftd}.

We use a standard BoW vector and its subtraction of two consecutive times in the FTD. First, a dictionary is generated at the StyleNet+L*a*b descriptor. The dictionary is calculated with the k-means algorithm from 1,600,000 randomly selected feature vectors in the FCDB. We set the number of codewords to 1,000 ($= K$), which has been given by good experimental results for modeling fashion style. In test time, a feature vector is assigned to the closest centroid in the dictionary for constructing a codeword vector.

The subtraction of codeword vectors is based on \cite{KataokaBMVC2016}, where positive $F^{+}$, negative $F^{-}$, and zero-mean value $F^{0}$ are equal to the subtraction value $\Delta v$ under the following three conditions:  
\begin{eqnarray}
\left\{
\begin{array}{l}
F_{i}^{+}(t) = |\Delta v^{t}_{i}| \;\; (\Delta v^{t}_{i} > TH) \\
F_{i}^{0}(t) = |\Delta v^{t}_{i}| \;\; (-TH < \Delta v^{t}_{i} < TH)\\
F_{i}^{-}(t) = |\Delta v^{t}_{i}| \;\; (\Delta v^{t}_{i} < -TH)
\end{array}
\right.
\end{eqnarray}
The subtraction of codeword vectors $\Delta v$ is calculated from the codeword vectors at time $t$ and $t-1$ ($\Delta v_{i}^{t} = v_{i}^{t} - v_{i}^{t-1}$). The codeword vector is $v \in \mathbb{R}^{1{,}000}$ at each word $i$, where $F^{+}$ and $F^{-}$ are the appearing and disappearing fashion styles, and $F^{0}$ shows ordinary fashion styles that are not changed at the time.

\section{Experimental results and discussion}

The paper applies the concept to worldwide fashion trend analyses.
We evaluate and visualize two different scenarios (spatial fashion analysis and temporal fashion analysis) on the FCDB. The spatial fashion analysis is executed to understand city perception and its similarity among 16 cities based on the conventional work~\cite{ZhouECCV2014}. The spatial fashion analysis is shown in~\ref{sec:cityperceptions} and~\ref{sec:citysimilarity}. Moreover, we verify the temporal fashion analysis in~\ref{sec:fashiontrend}. The temporal fashion analysis unveils worldwide fashion trends in a time sequence.  The representative fashion trends are clearly visualized in the experiment. We also show a feature improvement for a better analysis in~\ref{sec:featureimprovement}. The improvement is started by StyleNet. We additionally concatenate local features such as L*a*b, HOG~\cite{DalalCVPR2005}, and DeCAF~\cite{DonahueICML2014}.

\subsection{Feature Improvement}
\label{sec:featureimprovement}

To add sophistication to a codeword-based feature representation from fashion snaps, we consider an improvement feature that significantly represents a fashion style in this subsection. We apply several features such as L*a*b (Lab) and DeCAF in addition to StyleNet as a baseline. That is, we answer whether StyleNet is enough as a fashion style descriptor. In the task of fashion style classification, we apply the HipsterWars dataset, which includes clear labels about fashion styles~\footnote{The Fashion144K dataset also has style labels but it has broad-sense attributes based on part-based labels}. We classify five fashion styles (Bohemian, Hipster, Pinup, Preppy, Goth) in  the Hipster Wars dataset.

The FCDB includes images that are taken under various difficulties such as illumination change, outdoor/indoor, and cluttered backgrounds. Therefore, we must consider combining a sophisticated feature with StyleNet as the objective in the paper. Here, Lab, AlexNet (AN), VGGNet (VN), HOG, and combined features are extracted from an input image. All feature dimensions are standardized as 128 dimensions with PCA. We compare original images and cropped-person images by R-CNN in the classification. The settings of SVM  are the RBF-kernel, $C=0.01$ and the train/test follows the HipsterWars dataset.

\begin{figure}[t]
       \centering
       \includegraphics[width=1.1\linewidth]{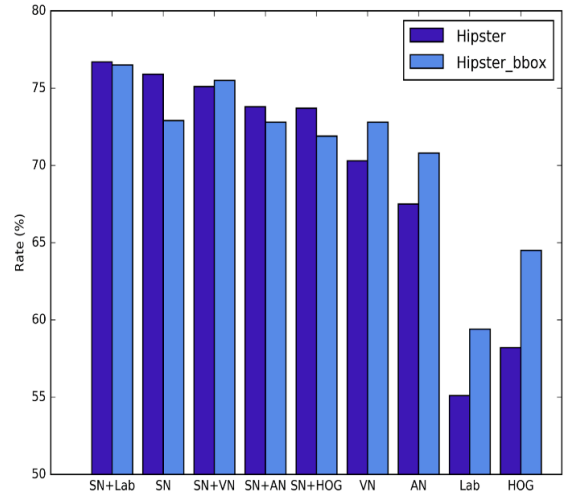}
        \caption{Comparison of feature representation}
       \label{fig:hipsterwars}
\end{figure}

The classification accuracy on HipsterWars is shown in Figure~\ref{fig:hipsterwars}. The figure gives the comparison results of an original style classification (Hipster) and a cropped-person style classification (Hipster\_bbox), i.e., not only the feature improvement from the baseline. The detected person scenario (Hipster\_bbox) corresponds to the FCDB, which crops persons by the detector. The StyleNet+Lab (SN+Lab) improvement is +0.8\% (Hipster) and +3.6\% (Hipster\_bbox) by comparing StyleNet (SN) only; namely, the improvement is shown by using the Lab color feature. The other combined features (e.g., SN+AN: 73.8\% and 72.8\%, SN+VN: 75.1\% and 75.5\%) did not reach the same improvement from the baseline. According to the results, the color information is one of the important elements in fashion style. In other words, the fashion style descriptor is complemented by the color feature. Therefore, we aim for an improvement feature that adds color information into the conventional fashion-oriented feature~\cite{SimoSerraCVPR2016}. 

\begin{figure}[t]
\centering
\subfigure[Codeword vector with StyleNet]{\includegraphics[width=0.48\linewidth]{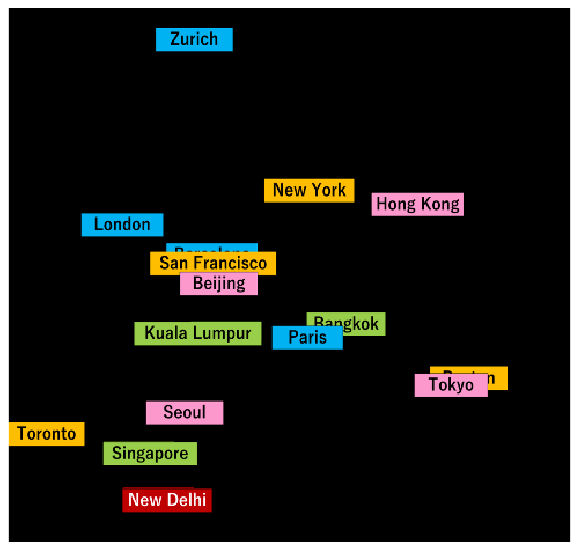}
\label{fig:tsne_st}}
\subfigure[Codeword vector with StyleNet+Lab]{\includegraphics[width=0.48\linewidth]{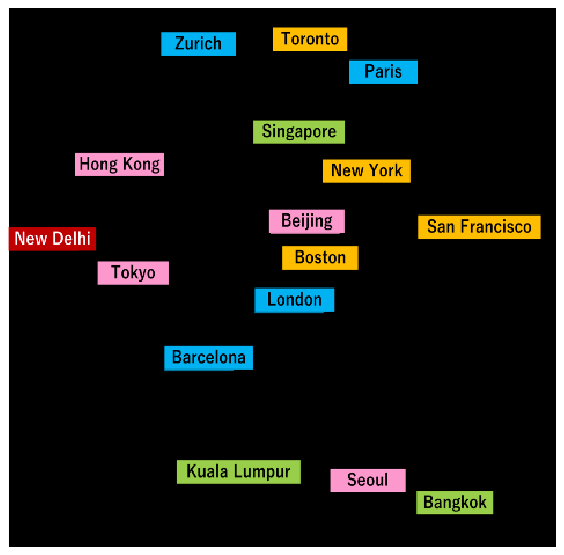}
\label{fig:tsne_stlab}}
\caption{Visualization with t-sne: (blue) Europe, (pink) East Asia, (green) Southeast Asia, (yellow) North America, (red) South Asia}
\label{fig:tsne_codeword}
\end{figure}

By comparing a codeword vector with the StyleNet and StyleNet+Lab, the visualization of feature improvement is shown in Figure~\ref{fig:tsne_codeword}. To discuss the effect of codeword performance in \ref{sec:cityperceptions} and \ref{sec:citysimilarity}, we show our feature improvement in the figure. We visualize the codeword vector for each city in Figure~\ref{fig:tsne_st} (only StyleNet) and Figure~\ref{fig:tsne_stlab} (StyleNet+Lab). We mapped these codeword vectors positioned by t-SNE ($t$=2).  Figure~\ref{fig:tsne_stlab} illustrates that the feature improvement clearly affects the distance among cities (the separability performance is better). 

\begin{figure}[t] 
\centering
 \includegraphics[width=1.0\linewidth]{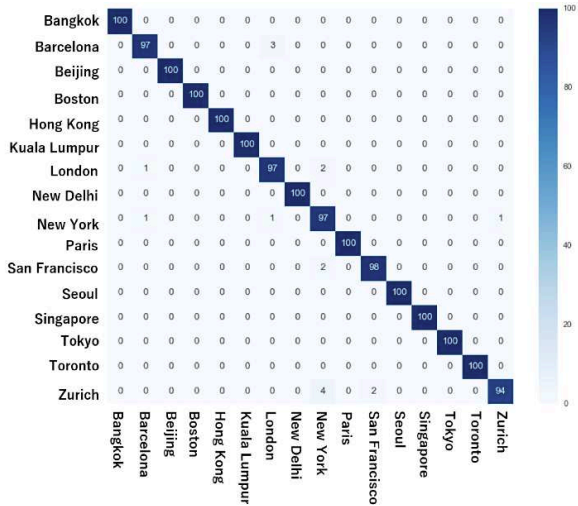}
 \caption{Confusion matrix from fashion-based city identity recognition with codeword vector}
\label{fig:confusion_st}
\end{figure}

\subsection{16-city perception}
\label{sec:cityperceptions}
To confirm the city identity based on previous work~\cite{ZhouECCV2014}, we classify between cities with a fashion-based codeword. We classify 16 cities in the FCDB with the codeword vectors of the combined StyleNet+Lab and BoW. These codeword vectors are generated by the fashion style descriptor extracted from city images and k-means clustering ($K$=1,000). Based on the previous works, we generate 500 training and 100 testing 1,000-dim codeword vectors. To make train/test codeword vectors for each city, we divide images from the FCDB. We randomly select 10,000 images and generate a train/test codeword vector. 

Figure~\ref{fig:confusion_st} shows the confusion matrix of the 16-city perception. The performance of city perception is a total of 98.9\%. The result suggests that a fashion-based codeword vector expresses the city characteristics. We also confirmed that a different culture tends to be a more discriminative feature (e.g., New York is more similar to other Western countries than to Asian countries). The results show that the fashion-based codeword is more discriminative than the scene-based feature: that is, people reflect  the city.

\subsection{City similarity}
\label{sec:citysimilarity}

\begin{figure}[t] 
       \centering
       \includegraphics[width=1.0\linewidth]{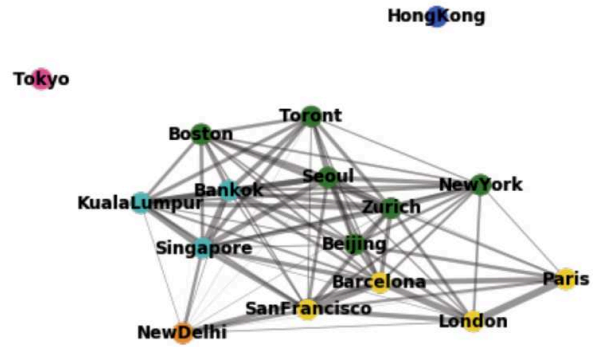}
        \caption{Fashion-based city similarity graph: The input is from 1,000-dim codeword vectors with StyleNet+Lab. The nodes and edges correspond to cities and similarities between pairs of cities, where the line thickness indicates the degree of similarity. In the graph representation, we set the thresholding value as 0.2 in order to eliminate low correlations.}
       \label{fig:city_sim_graph}
\end{figure}

We show the similarity or difference between two cities by using the fashion style descriptor. In the city similarity, our approach basically aligns with Zhou's analysis of scene-based city similarity~\cite{ZhouECCV2014} in the graph representation. Figure~\ref{fig:city_sim_graph} illustrates the fashion-based city similarity by referring to the scene-based city similarity. 

According to Figure~\ref{fig:confusion_st}, the perfectly percepted  cities (e.g., Paris, Kuala Lumpur, and Tokyo) tend to have separatable  characteristics in the codeword vector, such as cultural background and climate. The knowledge is similar to the fashion-specified ``What Makes Paris Look like Paris?" in 2012~\cite{DoerschToG2012}. The cities that do not have 100\% perception are seen in the European countries (Barcelona, London, and Zurich, which are connected by land) and the US (New York and San Francisco, which are in the same country). The two tendencies come from similar backgrounds such as language, climate, and short distance. These fashionabilities are inevitably standardized against each other.

Cities such as Paris, San Francisco, and London have strong similarities to each other. These are so-called fashionable cities; however, their label is not a sufficient reason to describe the similarities. One reason is described by the World Tourism Barometer~\cite{tourismrank}. The tourism-rank statistics for 2013--2015 show France \& the US are the top 2, and the UK is 8th. Especially, Paris, San Francisco, and London are dense tourist areas. Although the averages of fashionabilities are affected by tourists, the fashion styles from traditional cultures allow us to highly classify between cities (see Figure~\ref{fig:confusion_st}).

The separability of Tokyo and Hong Kong is high; namely, the two countries have strong uniqueness. 
Especially, Tokyo has fashion-specific characteristics in our analysis. Recently, different fashion cultures appear at each area in Tokyo. The cultures are coming from traditional \& latest cultures. Another tendency shows social activities from history such as Singapore, Bankok and Kuala Lumpur.

The knowledge of the fashion-based city similarity is different from the scene-based one, because geographical restrictions are not subject to the recent developments of the World Wide Web and transportation networks. 


\begin{figure*}[t]
\centering
 \includegraphics[width=0.95\linewidth]{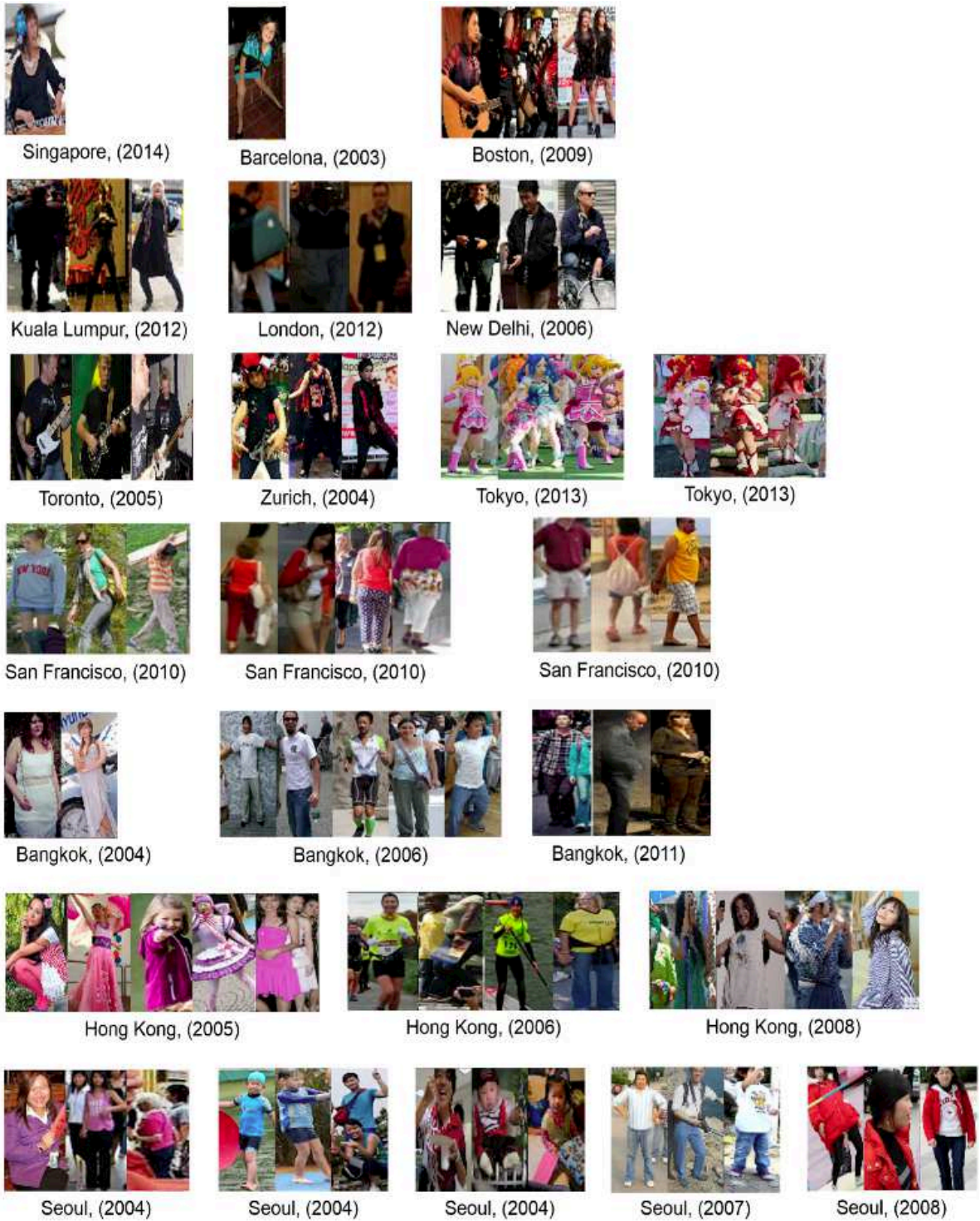}
 \caption{Visualization of fashion trends: The figure shows randomly selected cutting-edge fashion trends per city and year. For example, Tokyo (2013) shows one of the cutting-edge fashion trend at Tokyo in 2013. Although the FTD is not perfect, we confirm the concept of ``changing fashion culture" is achieved as the result of FCDB collection and unsupervised analyzer.}
\label{fig:ftd_visualization}
\end{figure*}

\subsection{Fashion trends}
\label{sec:fashiontrend}

In this section we show the worldwide fashion trends at each year relative to the previous year.  We assign our FTD as an unsupervised analyzer to clearly and explicitly depict the fashion trends as a result of temporal subtraction.

Here we evaluate fashionability changes per city with our FTD. The codeword dictionary is generated with the images over 16 years (from 2000 to 2015) from the FCDB. The FTD, especially parameters $F^{+}, F^{-}$, visualizes the fashionability changes by the values of the differences ($>TH=0.01$, where the value is experimentally decided) at each codeword. We take a couple of representative fashion snaps that are the nearest samples~\footnote{Randomly selected nearest images} to centroids. We believe that the nearest fashion snaps represent a strong fashionability at the codeword. Figure~\ref{fig:ftd_visualization} illustrates the changing fashion trends in the randomly selected cities and years. 


\section{Conclusion}
\label{sec:conlusion}
In the paper, we proposed a novel fashion trend analysis from a large-scale database collection and an unsupervised framework. We collected the fashion culture database (FCDB), which consists of 76 million geo-tagged images in 16 cosmopolitan cities. To visualize the cutting-edge fashion trends, we also proposed an unsupervised fashion trend descriptor (FTD) that is composed by a fashion style descriptor, a codeword vector, and temporal subtraction. The combination of our FCDB and FTD significantly visualizes worldwide fashion trends in a time series. In the future, we plan to collect fashion snaps over a longer timeframe (e.g., 30 years) in order to observe the periodic changes of fashion cultures in different countries.

{\small
\bibliographystyle{ieee}
\bibliography{fashionculture}
}

\end{document}